# An Objectness Score for Accurate and Fast Detection during Navigation

*Hongsun Choi[1], Mincheul Kang[2], Youngsun Kwon[3] and Sung-eui Yoon[4]

[1] *Software Graduate Program, KAIST, Seoul 06301, Korea*
[2], [3], [4] *School of Computing, KAIST, Daejeon 34141, Korea*

[1] hs.choi@kaist.ac.kr, [2] mincheul.kang@kaist.ac.kr, [3] youngsun.kwon@kaist.ac.kr, [4] sungeui@kaist.edu

## ABSTRACT

We propose a novel method utilizing an objectness score for maintaining the locations and classes of objects detected from Mask R-CNN during mobile robot navigation. The objectness score is defined to measure how well the detector identifies the locations and classes of objects during navigation. Specifically, it is designed to increase when there is sufficient distance between a detected object and the camera. During the navigation process, we transform the locations of objects in 3D world coordinates into 2D image coordinates through an affine projection and decide whether to retain the classes of detected objects using the objectness score. We conducted experiments to determine how well the locations and classes of detected objects are maintained at various angles and positions. Experimental results showed that our approach is efficient and robust, regardless of changing angles and distances.

## 1. INTRODUCTION

Recently, many applications involving deep learning networks (DLN) have been developed in various fields. Specifically, computer-vision-based DLNs such as Faster R-CNN (Ren 2015), Mask R-CNN (He 2017), and YOLOv3 (Redmon 2018) are notable examples that are widely utilized in the field of robotics for recognizing objects, obstacles, and terrain.

These networks, however, were not designed directly for robotic applications. In conventional robotic applications for navigation, they used not only a real-time grid-based occupancy mapping (Kwon 2019) but also a simultaneous localization and mapping (SLAM) to track landmarks extracted from features using an image retrieval method (Kim 2018). On the other hand, in order to increase more accuracy of conventional algorithms, some works have been proposed to utilize semantic information of objects detected from

---

[1) 2) 3)] Graduate Student
[4)] Professor

DLN-based object detection modules. Pirsiavash et al. (2011) proposed an algorithm for multi-object tracking, including estimating the number of objects and tracking their states. However, this method tracks moving objects without camera movement. Bowman et al. (2017) used detected objects as semantic information for semantic SLAM. When an object in the same position is detected as belonging to different classes, the class which has the highest score from the object detector is used as semantic information. If the shape of the object changes on the 2D image plane during navigation, the object detector often misinterprets the class of the detected object.

As a result, these prior approaches did not consider which objects were previously detected during navigation, and thus often misinterpreted objects, especially, when a robot approaches an object closely (Fig. 1). For example, when a mobile robot with a forward-looking camera moves, an object seen on a 2D view image plane at a certain location undergoes complex transformations including rotations, affine transforms, and scale changes in different view planes seen from other positions, causing the robot to misinterpret the locations and classes of objects during navigation.

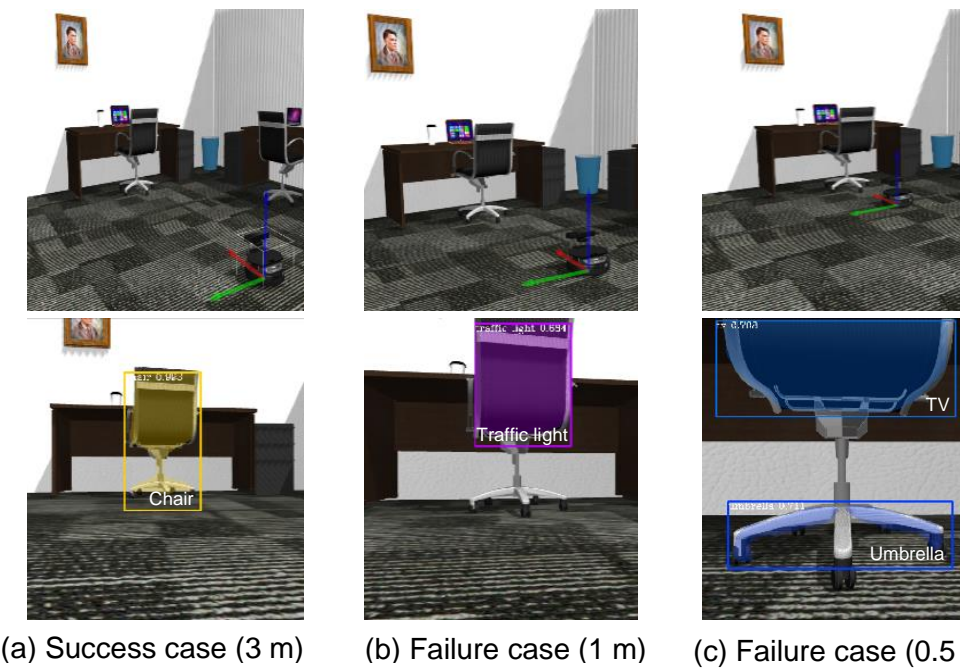

(a) Success case (3 m)   (b) Failure case (1 m)   (c) Failure case (0.5 m)

Fig. 1 An example case of the scale change problem. As the robot gets closer, the detected classes are changed. (a) The successful case at 3 m. (b) The chair is misinterpreted as a traffic light at 1 m and then (c) is misinterpreted as a television and an umbrella at 0.3 m.

In this paper, we propose a novel method to maintain the locations and classes of objects detected from an object detection module (e.g., Mask R-CNN). To achieve our goal, we use the concept of an objectness score based on the assumption that we can detect objects with high accuracy at an appropriate distance, not too far away nor too close to the object.

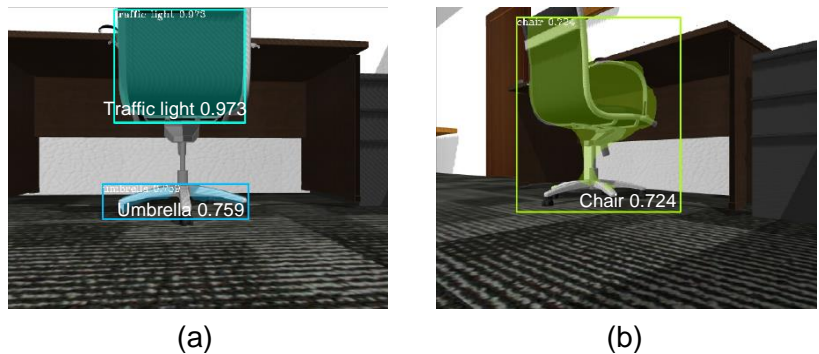

(a)             (b)

Fig. 2 These images show how different classes may be assigned to the same object (chair) from Mask R-CNN. (a) This image shows that the detector estimates the upper part of the chair as a traffic light with a probability of 0.973 and the lower part of the chair as an umbrella (0.759). (b) This image shows a case in which the detector predicts the chair (0.724) successfully; the numbers in parentheses indicate the class probability. If we use only the class probability to determine the class, we can then misinterpret the chair as a traffic light.

## 2. ACCURATE DETECTION DURING NAVIGATION

Object detection modules (e.g., YOLO v3, Mask R-CNN, etc.) can estimate the classes and bounding boxes of objects on a 2D image plane in every frame. As shown in Fig. 2, the 2D view image plane can undergo rotations, affine transformations, and scale changes. These issues pose a challenge to detection modules such that the mobile robot may misinterpret the class of an object in one frame, while they work well to identify the same object in other frames.

To address this problem, the state-of-the-art methods (Song 2016 and Qi 2018) utilize a 3D intersection over union (IOU) and class probabilities between the 3D bounding boxes of detected objects. However, we found that these methods can require high amounts of computational overhead (e.g., 120 ms) to estimate a 3D bounding box and calculate the 3D IOU.

In this paper, we use the 3D positions and the 2D IOU of detected objects for efficient computation instead of the 3D IOU. We first discuss how to maintain and compare the 2D bounding boxes of detected objects in every frame in Section 2.1. Section 2.2 introduces an objectness score for determining the exact class among the different detected classes for an object in the same position.

### 2.1. MAINTAINING THE LOCATIONS AND CLASSES OF OBJECTS DETECTED

Most mobile robots using a 3D object detection module try to map the 3D bounding boxes of objects detected during navigation, and then utilize that information when the mobile robot comes back to that region. While the 3D object detection modules require high computational power, their performance could be lower than that of 2D detection modules due to the higher complexity of 3D mapping. For more accuracy and efficient computation, we propose to use a 2D object detection module and 2D IOU.

As shown in Fig. 3, our method consists of three parts. First of all, we use a 2D

object detection module (e.g., Mask R-CNN) to recognize the 2D bounding boxes and class probabilities of any objects detected. Second, we convert the 2D bounding box to a 3D location using the mean depth of a point cloud set corresponding to the 2D bounding box region and record the class and the 3D location of the detected object in 3D world coordinates to use this information during navigation. Lastly, if the mobile robot detects the same object as a different class from the previously detected class, we compare the previous class with the current one and use the objectness score to select the more accurate class label.

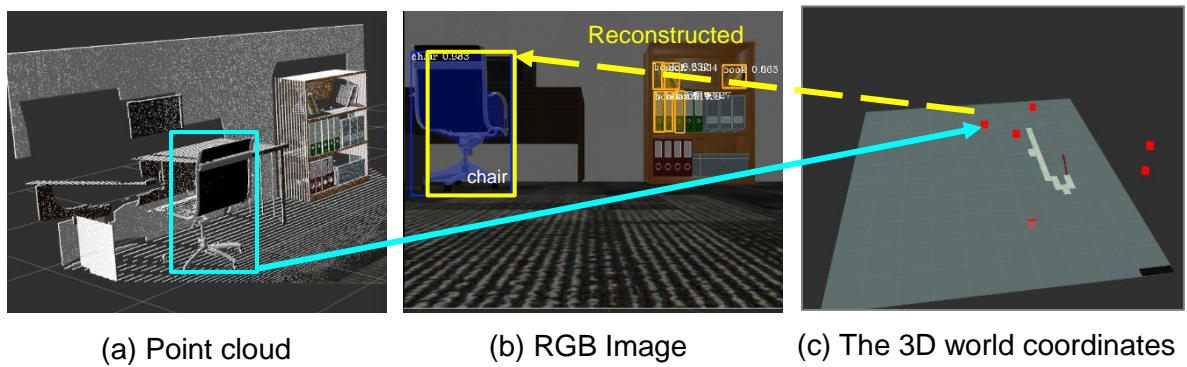

(a) Point cloud     (b) RGB Image     (c) The 3D world coordinates

Fig. 3 These figures show how our approach maintains the locations and classes of detected objects in the 3D space, and reconstructs them onto the 2D image plane. The images in (a) and (b) show a point cloud and RGB image from the RGB-D camera mounted on a mobile robot. (c) is a reconstructed 3D map in the 3D world coordinates. When the robot recognizes a chair during navigation, we obtain the point cloud set in the 2D bounding box region of the chair (the box in (a)) and convert the point cloud of the chair to a 3D location (the red point in (c)). This 3D information contains the 2D bounding box location, class probability, mean depth, etc. When the chair is observed in the field of view again, we then project the 3D location of the previously detected object onto the RGB image plane (the yellow box in (b)) using affine projection. Based on this information, we can compare which class is more accurate between the reconstructed 2D bounding box (the yellow box in (b)) and the currently detected 2D bounding box (the blue box in (b)).

To compare between different classes in the same location, we reconstruct the 2D bounding box from the saved 3D object information using an affine projection. In homogeneous coordinates, the affine projection is represented as the following:

$$p_{img} = K[R|t]P_{world}, \qquad (1)$$

where $p_{img}$ is $[x, y, 1]^T$ in the 2D image coordinates, $P_{world}$ is $[X, Y, Z, 1]^T$ in the 3D world coordinates, K is an intrinsic matrix, and [R|t] is an extrinsic matrix. We obtain the intrinsic matrix from camera calibration and the extrinsic matrix from odometry of the mobile robot. Thus, we can save information about detected objects such as their location and class regardless of the motion of the mobile robot.

**2.2. CLASS COMPARISON USING AN OBJECTNESS SCORE**

In this section, we introduce a novel method to determine which class to choose when an object is detected as belonging to different classes. Most DLNs train their network architectures with an image dataset containing the overall shapes of objects. Therefore, it is necessary to maintain a sufficient distance between the robot and objects for DLNs to perform accurate object detection. Using this property, we formulate an objectness score ($S_{obj}$) and define an object managing algorithm (Alg. 1). $S_{obj}$ is used to determine which object class is more accurate when one object is recognized as belonging to multiple classes. Alg. 1 computes the 2D IOU to determine whether it is the same object and uses the objectness score to select a more accurate class label.

Computer-vision-based DLNs such as Faster R-CNN (Ren 2015), Mask R-CNN (He 2017), and YOLOv3 (Redmon 2018) estimate 2D bounding boxes and classes which have class probabilities. Traditional methods (e.g., Bowman 2017 and Qi 2018) use class probabilities from DLNs to determine which class assignment to keep. Class probabilities are very simple and useful, but those often make a mistake when only a part of the object is observed along the distance. We assume that an estimated class is more accurate when the DLNs can see the entire shape of an object. Therefore, considering the distance from the object, we define an objectness score ($S_{obj}$) which is designed to give more weight to observations made when the distance is sufficient for the whole object shape to be observed. Here, S and obj mean a score and an object, respectively.

**Objectness Score ($S_{obj}$).** The objectness score is defined as the following:

$$S_{obj} = \alpha S_{poc} + (1-\alpha) S_{\text{depth}}, \tag{3}$$

where $S_{poc}$ is a class probability estimated by a DLN such as Mask R-CNN, $S_{\text{depth}}$ is the normalized distance score between the camera and an object, and α is a relative weight between $S_{poc}$ and $S_{\text{depth}}$. Intuitively speaking, we give a higher score when we have a high class probability from detecting an object that is also located a sufficient distance from the camera.

In more detail, if the distance between the robot and an object is too great, the object looks like a dot. This problem causes DLNs to misinterpret the class of the detected object. Thus, we define $S_{\text{depth}}$ by using field-of-view information from the RGB-D camera to quantify a sufficient distance for recognizing the entire shape of the object. To realize this idea, we define $S_{\text{depth}}$ as a normalized distance with a value between 0 and 1 using min-max normalization:

$$S_{\text{depth}} = \frac{d_i - \min(d)}{\max(d) - \min(d)}, \tag{2}$$

where $d_i$ is the distance between a robot and the i-th detected object, and $\min(d)$ and $\max(d)$ are pre-defined values for the minimum and maximum distance from the RGB-D camera mounted on the robot for observing objects in a sufficient distance range; in our test, $\min(d)$ and $\max(d)$ are set to be 0.8 m and 3 m respectively.

We now explain how our algorithm (Alg. 1) works. Alg. 1 selects a more accurate object through the objectness score and stores the detected object information such as a 3D position, 2D bounding box, class, and objectness score in the list of detected objects

($P_{obj}$) during navigation. When the list of detected objects $P_{obj}$ is empty and an object is detected, the detected object is stored in $P_{obj}$ (Line: 1). Then, if a new object $o_{new}$ is detected, we search the k-nearest neighbor objects ($\rho_1$, $\rho_2$, …, $\rho_k$) of $o_{new}$ from $P_{obj}$ using *SearchKNN(·)* (Line: 6), which returns the k-nearest neighbor objects (k=3). Subsequently, we check whether $o_{new}$ and any of the k-nearest neighbor objects share the same position through *Check2DIOU(·)* (Line: 8), which returns *true* if the 2D IOU (intersection over union) between $o_{new}$ and $\rho_i$ is larger than a threshold value (0.9), where $\rho_i$ is the i-th nearest neighbor object. If *Check2DIOU(·)* is *true*, we consider $o_{new}$ and $\rho_i$ to be in the same location. Thus, if the objectness score of $o_{new}$ ($S_{o_{new}}$) is larger than the objectness score of the existing object $\rho_i$ ($S_{\rho_i}$), $\rho_i$ is replaced by $o_{new}$ because $o_{new}$ has more accurate object information than $\rho_i$ according to the objectness score.

In addition, if $o_{new}$ does not overlap with any one of the k-nearest neighbor objects ($\rho_1$, $\rho_2$, …, $\rho_k$), $o_{new}$ is newly inserted in $P_{obj}$ (Line: 12, 13).

---
**Algorithm 1** Class selection algorithm
---
**Input:** A newly detected object $o_{new}$,
  A list of detected objects $P_{obj}$
**Output:** Updated $P_{obj}$
  1: **if** $P_{obj}$ is empty **then**
  2:   Insert $o_{new}$ into $P_{obj}$;
  3: **end if**
  4: $\rho_i \leftarrow$ *SearchKNN($o_{new}$)*;
  5: *Calculate* an objectness score of $o_{new}$, $S_{o_{new}}$;
  6: **while** $\rho_i$ is exist **do**:
  7:   *Calculate* an objectness score of $\rho_i$, $S_{\rho_i}$;
  8:   **if** *Check2DIOU($o_{new}$, $\rho_i$)* **and** $S_{o_{new}}$ is larger than $S_{\rho_i}$ **then**
  9:     *Replace* $\rho_i$ with $o_{new}$;
 10:   **end if**
 11: **end while**
 12: **if** $o_{new}$ is not the same location as all $\rho_i$ **then**
 13:   *Insert* $o_{new}$ into $P_{obj}$;
 14: **end if**

## 3. EXPERIMENTS

The goal of our experiment is to maintain the locations and classes of detected objects during navigation. To verify our system, we experimented in the Gazebo simulator within the Robot Operating System (ROS). We used a mobile robot based on a turtlebot model with a mounted RGB-D Kinect camera and the mobile robot moved in the indoor environment (Rasouli 2017). We set the weight $\alpha = 0.4$ in the objectness score shown in Eq. (3). Intrinsic parameters were obtained using a camera calibration technique (Zhang 2000, Khoshelham 2012) and extrinsic parameters were obtained from the pose of the camera mounted on the mobile robot. We tested our algorithm by changing the angle (-45, 0, and 45 degrees) of the mobile robot and the distance (from 0.3 to 3 m). The minimum and maximum distances of $S_{\text{depth}}$ were defined as 0.8 m and 3 m, respectively. The experimental results (Fig. 3) showed that the location and class of the detected object were maintained when the mobile robot moved toward the object from different angles and distances.

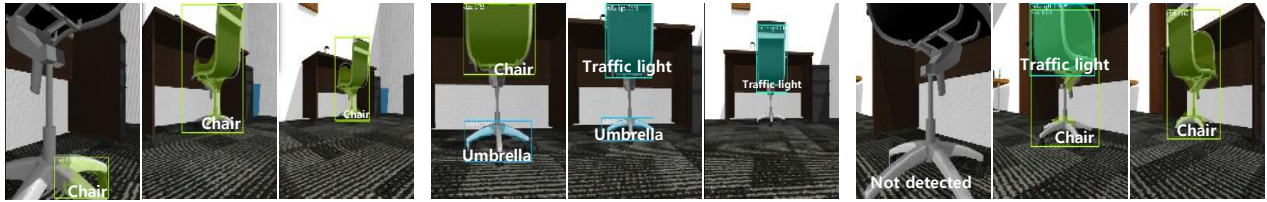

(a) The results of classification with localization from Mask R-CNN

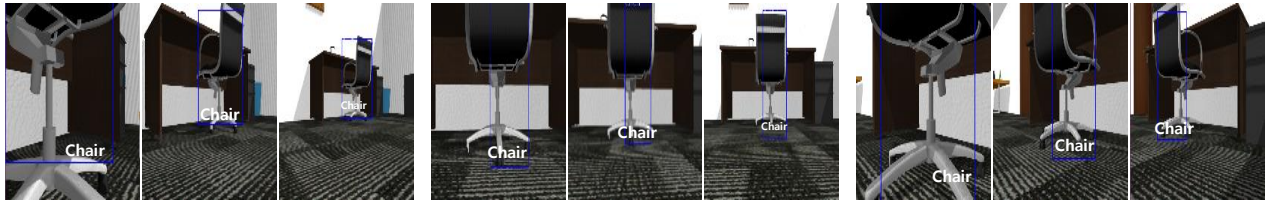

(b) The results of our algorithm

Fig. 4 These image sequences show the results of classification with localization during navigation of a mobile robot. Our method robustly identifies objects during navigation.

## 4. CONCLUSION AND FUTURE WORK

In this paper, we have proposed a method for maintaining the locations and classes of detected objects using the objectness score. The objectness score determines which objects to maintain between currently and previously detected objects. Experimental results at various angles and distances showed that our method was robust for detecting objects during navigation. In future work, we will extend this approach to distinguish whether or not the maintained object is an obstacle that we need to avoid during navigation.

## ACKNOWLEDGMENT

This research was supported by the SW Starlab support program (IITP-2015-0-00199) and the Ministry of Trade, Industry & Energy (MOTIE, Korea). No.10067202